\documentclass[9pt,technote]{IEEEtran}

\usepackage{times} 
\usepackage[T1]{fontenc}         
\usepackage{amsmath}
\usepackage{bm}
\usepackage{multicol}
\usepackage{draftwatermark}
\SetWatermarkText{PRE-PRINT}
\SetWatermarkLightness{.85}
\SetWatermarkScale{.8}

\begin{document}


\title{Neural embeddings for metaphor detection in a corpus of Greek texts}

\author{\IEEEauthorblockN{
Eirini Florou\IEEEauthorrefmark{1},Konstantinos Perifanos\IEEEauthorrefmark{2}, Dionysis Goutsos\IEEEauthorrefmark{3}}\\
\IEEEauthorblockA{ Department of Linguistics, National and Kapodistrian University of Athens
\\
\IEEEauthorrefmark{1}eirini.florou@gmail.com,
\IEEEauthorrefmark{2}kostas.perifanos@gmail.com,
\IEEEauthorrefmark{3}dgoutsos@phil.uoa.gr}}

\date{}

\maketitle
 
\begin{abstract}
\noindent
One of the major challenges that NLP faces is metaphor detection, especially by automatic means, a task that becomes even more difficult for languages lacking in linguistic resources and tools. Our purpose is the automatic differentiation between literal and metaphorical meaning in authentic non-annotated phrases from the Corpus of Greek Texts by means of computational methods of machine learning. For this purpose the theoretical background of distributional semantics is discussed and employed. Distributional Semantics Theory develops concepts and methods for the quantification and classification of semantic similarities displayed by linguistic elements in large amounts of linguistic data according to their distributional properties. In accordance with this model, the approach followed in the thesis takes into account the linguistic context for the computation of the distributional representation of phrases in geometrical space, as well as for their comparison with the distributional representations of other phrases, whose function in speech is already "known" with the objective to reach conclusions about their literal or metaphorical function in the specific linguistic context. This procedure aims at dealing with the lack of linguistic resources for the Greek language, as the almost impossible up to now semantic comparison between "phrases", takes the form of an arithmetical comparison of their distributional representations in geometrical space.

\end{abstract} 

\thispagestyle{empty}

\section{Introduction}
The recognition of the omnipresence and importance of metaphor in everyday communication, along with the difficulties of clearly defining it, accounts for the plethora of NLP systems that aim at the automatic differentiation between literal and metaphorical meaning. The vast majority of these systems try to detect metaphor based on selectional preference violation, ontologies, word taxonomies, statistical models or supervised machine learning \cite{metaphor_computational}. Although these approaches may be sufficient for resource-rich language, they work less well for languages lacking in such resources. This paper uses the theoretical background of distributional semantics for the development of a neural language model employed for the automatic differentiation between literal and metaphorical meaning in authentic non-annotated phrases from a corpus of Greek.
Most theoretical approaches to metaphor, including Lakoff and Johnson’s \cite{lakov}  cognitive approach, conceive it as a means of mapping and connecting dissimilar conceptual areas. Treating metaphor as a concept and relating it to cognition, albeit useful, makes it necessary to employ semantic rules or features for an automatic recognition of metaphorical meanings and their differentiation from literal ones. Thus, the majority of computational models for the automatic detection and recognition of metaphors requires access to linguistic resources and tools or expensive and time-consuming manual annotation in order to identify semantic mappings. Taking into consideration the limitations of Greek as regards resources and tools, a different approach is suggested here, following the principles of distributional semantics that entail calculating the relations of a word with its linguistic context without explicitly taking into account any connections between this word and its related concepts. Our research explores the usefulness of neural language models in metaphor detection by employing a variety of word embedding algorithms. Here we discuss two of those, namely the neural log-linear model word2vec, using the continuous bag of words method (CBOW) \cite{mikolov2013distributed} and the co-occurrence matrix factorization model GloVe \cite{pennington2014glove}, in order to learn word embeddings.

\section{Computational approaches to metaphor}

The attempt to computationally identify and interpret metaphors has been initially based on violations of selectional preferences (e.g. Wilks \cite{wilks}; see also Shutova \cite{shutova_2}) or word taxonomies \cite{met_probabilities}; Xing et al. \cite{zhang}, involving rich semantic knowledge. Metaphor research has progressively moved towards the supervised learning of metaphor through the use of statistical models (\cite{met_dect_term_relev}), clustering (\cite{Birke}), Latent Dirichlet Allocation, LDA \cite{blei2003latent} or logistic regression \cite{metaphor_id}. 
As has happened with many linguistic phenomena, computational approaches to metaphor are now increasingly based on neural models. For instance, Mohler et al. \cite{mohler_metaphor} have used support vector machines (SVMs) to assign conceptual metaphors to metaphorical expressions. Their experiments involve three target domains and compile a predefined set of source domains with the target with which they are typically associated. Mohler et al. extract vectors for source domain words in metaphors and vectors of the given source concepts and then estimate their similarity as objects in geometrical space. They experiment with three types of models and compare their results with a human-annotated gold standard of source domain assignments. The Latent Semantic Analysis (LSA) model has a score of 0.54 in accuracy in English, while the neural network, which was based on word2vec for the identification of word embeddings, has higher accuracy in Spanish (0.59) and lower in Russian (0.49) and Farsi (0.48).
Shutova et al. \cite{shutova} and Bollegala and Shutova \cite{bollegala2013metaphor} have used unsupervised learning to paraphrase metaphor. Shutova et al. \cite{shutova} have used a vector space model in order to estimate word embeddings and then compute candidate paraphrases according to the context in which a metaphor appears. The meaning of a word instance in context is computed by adapting its original vector representation to the dependency relations in which it participates. For this purpose, this approach builds a factorization model in which words, together with the other words with which they share a context window and their dependency relations, are linked to latent dimensions \cite{vandecruys}.  A SP model is used to measure the degree of literalness of paraphrases. This method has shown a top-rank precision of 0.52. Bollegala and Shutova \cite{bollegala2013metaphor} have used a similar method in a set of candidate paraphrases from the World Wide Web, reporting a precision score of 0.42 for this model.

\section{Developing a neural network for identifying metaphor in Greek}

Distributional Semantics Theory develops concepts and methods for the quantification and classification of semantic similarities displayed by linguistic elements in large amounts of linguistic data according to their distributional properties \cite{harris_distributional}. Taking into account the linguistic context for the computation of the distributional representation of phrases in geometrical space, as well as for their comparison with the distributional representations of other phrases, whose function in speech is already "known", the objective is to reach conclusions about their literal or metaphorical function in the specific linguistic context. This procedure aims at dealing with the lack of linguistic resources for Greek, as the almost impossible up to now semantic comparison between phrases takes the form of a numerical comparison of their distributional representations in geometrical space. Distributional semantics are thus paramount in shifting research interest towards neural language models, which can attribute hidden statistical characteristics of the distributional representations of word sequences in natural language. 

Neural language models are preferred to traditional statistical models, as they are in position to render a large amount of training data with a relatively small number of parameters. Thus, the goal is to match every word with a constantly evaluated distributional representation in geometrical space, since the probability of a lexical sequence is expressed as the result of the appearance probabilities, under certain conditions, of the next word offering the probabilities of previous ones. Bengio et al. \cite{bengio} have proposed to fight the curse of dimensionality by learning a distributed representation for words, which allows each training sentence to inform the model about an exponential number of semantically neighboring sentences. The model learns simultaneously a distributed representation for each word, along with the probability function for word sequences, expressed in terms of these representations.

In order to achieve the matching of linguistic data with their distributional representations in geometrical space, the models word2vec by Mikolov et al. \cite{mikolov2013distributed} and GloVe by Pennington et.al. \cite{pennington2014glove}  are employed in this paper. 

Both approaches taks advantage of the context of each term in order to discriminate its function and uses.

\subsection{word2vec}

Word2vec is a particularly computationally-efficient predictive model for learning word embeddings from raw text. The algorithm impements two achitecures, Skip-Gram where the model tries to predict the word given the context and CBOW, where the model tries to predict the next word given its context.

In word2vec/CBOW used in this study, the neural net learns word representations by maximize the negative log-likelihood objective function

	$$ J  = -logP( w_c | w_{c-m}, ..., w_{c-1}, w_{c+1}, ...., w_{c+m} ) $$ 
	 where $c$, the context window size, $w_i$ the word $i$ of the vocabulary $V$.
	 
 The input of the word2vec neural net involves authentic and non-annotated texts and produces vectors as an output. The main purpose of an algorithm is to group vectors of similar words together in vector space. In this way, it converts words' semantic comparison into a mathematical computation of the distance between points in geometrical space. The quality of word vectors is determined by several factors, including the amount and quality of the training data and the size of vectors. Word2vec is an excellent technique for generating distributional similarity, since human tagged data are not needed; it is also fast to train, compared to previous techniques, and is able to manipulate either a small or a big amount of datasets.

\subsection{Glove}

A different approach is followed in GloVe, where word representations are learned by minimizing the objective function,

 	$$ J = \sum_{i,j=1}^{V} f(X_{ij}) \dot ( w_i^T \tilde{w_j} + b_i + \tilde{b_j} - log(X_{ij}))^2$$
	
	where $V$ the number of words in the vocabulary, $X_{ij}$ the number of appearances of words i,j in a pre-defined context window C (typically $C=10$),  $ w_i, \tilde{w_j}$ the word vectors,  $ b_i, \tilde{b_j}$ biases for words i, j, $f(x)$ a weight function to smooth out the effect of frequent words, 
	\[
  f(x) =   
  \begin{cases}
    (x/x_{max})^a, x < x_{max}  \\
    1, \quad  x \geq  x_{max}
  \end{cases}
\]

 $a = 3/4$, $x_{max} = 100$.

\subsection{Methodology}

For our experiments, training and testing data are taken from the Corpus of Greek Texts. The whole corpus (30 million words approx.) constitutes the input of the word2vec algorithm, which produces word vectors. We have chosen twenty four transitive verbs in Greek as used in 914 phrases including object complements. The distinction between literal and metaphor phrases has been based on the Metaphor Identification Procedure (MIP) suggested by the Pragglejaz Group \cite{steen2007finding}. By applying MIP to phrases from the Corpus of Greek Texts, we have manually created two lists, one with 459 literal sentences and another one with 455 metaphorical ones. In these two lists the same verbs are used as sentence predicates with a variety of different objects. Collocations and delexical verbs were excluded from the testing corpus. After word2vec produced vectors, a t-test was used in order to compare the means of the metaphor and the literal group which follow a normal distribution. The t-test assessed that the means of the two groups are statistically different from each other and thus we were able to use word embeddings in order to differentiate metaphor from literal cases. For this purpose, the linguistic context of the phrase was searched, since this is the factor which determines its use and function. An aggregate vector was thus created to represents each sentence and the label metaphor or literal was assigned as a consequence of the MIP. The classification of the support vector machine was based on these elements, i.e. the aggregate distributional representation of each sentence and the distinction between the literal and the metaphoric, according to this procedure.

\section{Results and discussion}

In order to evaluate model performance a 10-fold cross validation was applied. In our experiments word2ec/CBOW outperforms GloVe in accuracy as shown in table \ref{table:results}.

\begin{table}[h]
\caption{ Results }
\label{table:results}
\begin{center}
\begin{tabular}{ |c|c|c|c| } 
 \hline
 
 \bf{Model} & $\bf{D}$ & \bf{Accuracy} & \bf{Precision} \\ 
 \hline 
 word2vec  & 450 & 0.82  & 0.81 \\
 GloVe & 400  & 0.61  & 0.59 \\ 
 \hline
\end{tabular}
\end{center}
\end{table}

The high accuracy from word2vec is especially good for resource-poor languages, since the almost impossible up to now semantic comparison between phrases has been substituted for a numerical comparison of their distributional representations in geometrical space. Word embeddings have thus proved to be useful tools for the estimation of the semantic closeness between terms, and even entire phrases.

In sum, a discriminative model is recommended in our research for the distinction between the literal and the metaphorical function of a phrase. This model, through appropriate training, is able to identify the optimal separating hyper plane of a vector representation word combination and has the ability to generalize to new, unseen data. Thus, an attempt is made to automatically detect and recognize metaphorical meaning in an automatic and dynamic way through the numerical comparison of the distributional representations of phrases with those of others, already characterized in terms of their function.

Thanks to neural language models it is possible to distinguish between metaphorical and literal phrases by computational means and, at the same time, to recognize the function of a term in a specific linguistic context. In addition, neural language models and machine learning, in general, can contribute to the overcoming of difficulties associated with the precise identification of a metaphor, including such questions as what the metaphor unit is and whether there are several types of metaphor. Approaches like ours suggest that the linguistic context of a phrase and their distributional representations is all important in addressing these questions. Our approach can be seen as contributing to computational semantics, as it addresses a thorny semantic issue in a highly computational way, since words are transcribed into vectors and measurements and comparisons can now be made on them.

This is also the first computational metaphor approach based on a Greek corpus. As no other theoretical or computational research on metaphor has been carried out in Greek, it has been necessary to start from a bilateral distinction between metaphorical and literal phrases, although it is possible that pure metaphor resides at a continuum of integration, at the other end of which there is the pure literal. At the same time, it remains to be seen whether neural language models are able to discriminate pure metaphor from other kinds of figurative speech such as metonymy, synecdoche etc. 
As noted at the beginning this paper belongs to a larger project, aiming at exploring the usefulness of machine learning algorithms in metaphor detection. Apart from investigating other algorithms, extensions of our work also include investigating whether word embeddings are capable of deriving every type of metaphor and not just those involving a verb and its object complement, as well as experimenting with more data from Greek and data from other languages.

\bibliographystyle{ieeetr}
\nocite{*} 

\bibliography{metaphor}

\end{document}